\RequirePackage[l2tabu, orthodox]{nag}
\documentclass[pdflatex,sn-mathphys-num,iicol]{sn-jnl} 

\usepackage[T1]{fontenc}
\usepackage{latexsym}
\usepackage{amssymb}
\usepackage{amsmath}
\usepackage{amsthm}
\usepackage{booktabs}
\usepackage[inline]{enumitem}
\usepackage{siunitx}
\usepackage{graphicx}
\usepackage{color}
\usepackage{placeins}
\usepackage{caption}
\newcommand{\BibTeX}{B\kern-.05em{\sc i\kern-.025em b}\kern-.08em\TeX}
\def\mathsemicolon{;}
\def\ivBrack#1{\left\llbracket{#1}\right\rrbracket}
\newcommand{\eucdist}[2]{\left\|#1 - #2\right\|}
\def\classmembset#1{\mathcal{F}_{#1}}
\def\classmembfun#1{f_{#1}}
\def\card#1{\left|{#1}\right|}

\DeclareMathOperator{\simfun}{sim}
\def\ptFiguresDirectory#1{./#1}

\begin{document}

\title{A Compound Classification System Based on Fuzzy Relations Applied to the Noise-Tolerant Control of a Bionic Hand via EMG Signal Recognition}

\author*[1]{\fnm{Pawel} \sur{Trajdos}}\email{pawel.trajdos@pwr.edu.pl}%\orcid{0000-0002-4337-6847}%TODO no macro in this style for ORCID.

\author[1]{\fnm{Marek} \sur{Kurzynski}}\email{marek.kurzynski@pwr.edu.pl}%\orcid{0000-0002-0401-2725}

\affil*[1]{\orgdiv{Faculty of Information and Communication Technology}, \orgname{Wroclaw University of Science and Technology}, \orgaddress{\street{Wybrzeze Wyspianskiego 27}, \city{Wroclaw}, \postcode{50-370}, \country{Poland}}}

\abstract{
Modern anthropomorphic upper limb bioprostheses are typically controlled by electromyographic (EMG) biosignals  using a pattern recognition scheme. 
Unfortunately, there are many factors originating from the human source of objects to be classified and from the human-prosthesis interface that make it difficult to obtain an acceptable classification quality.
One of these factors is the high susceptibility of biosignals to contamination, which can considerably reduce the quality of classification of a recognition system.

In the paper, the authors propose a new recognition system intended for EMG based control of the hand prosthesis with detection of contaminated biosignals in order to mitigate the adverse effect of contaminations. The system consists of two ensembles: the set of one-class classifiers (OCC) to assess the degree of contamination of individual channels and the ensemble of K-nearest neighbours (KNN) classifier to recognise the patient's intent. 
For all recognition systems, an original, coherent fuzzy model was developed, which allows the use of a uniform soft (fuzzy) decision scheme throughout the recognition process.
The experimental evaluation was conducted using real biosignals from a public repository. The goal was to provide an experimental comparative analysis of the parameters and procedures of the developed method on which the quality of the recognition system depends. The proposed fuzzy recognition system was also compared with similar systems described in the literature.  
}
% %TODO up to 250 words! Now it has 218


% % \flushbottom %default
% \raggedbottom
% %%%%%%%%%%%%%%%%%%%%%%%%%%%%%%%%%%%%%%%%%%%%%%%%%%%%%%%%%%%%%%%%%%%%%%%%

\maketitle

\section{Introduction}\label{sec:Introduction}

Over the last two decades, there has been intensive development of methods for controlling upper limb prostheses based on biosignal recognition.
In this approach, the patient's intention to move the prosthesis is encoded in the contraction of the forearm residual muscles using the phantom motor execution (PME) phenomenon \cite{Garbarini2018}, through which the motor cortex directs movement of the forearm stump. Muscle activity generates electromyographic biosignals that can be recorded non-invasively from the body surface by registering the voltage between two electrodes (so called surface EMG - sEMG), distributed evenly around the stump~\cite{Witkowski2025}. The electronics of the measuring system (called the human-prosthesis interface - HPI) consists of sEMG sensors, filters, amplifiers, and A/D converters organised into a number of biosignal registration channels.
In the software implemented in the prosthesis, the recorded multi-channel sEMG signals are subject to subsequent steps of the feature engineering and classification procedures, which depend on the signal recognition method used.

Finally, the kinematic controller controls the degrees of freedom (DoF) of the prosthesis in such a way that the movement of the mechanical structure is consistent with the recognised class (type of movement).

Nowadays, constructing a multifunctional prosthetic device with dexterous mechanical capability and electric actuators, featuring multiple DoFs, is no longer an issue \cite{Calado2019}. The challenge is the imperfect prosthesis control system. It is still limited in terms of the range of movements (grips and manipulations) that the patient (human controller) can control and the correct operation of the prosthesis, which means performing the movement in accordance with the patient's intention. Both factors have a source in the imperfection of the recognition system.
The first factor is easy to identify, as it results from the limitation of the human controller or -- from a recognition perspective -- the human source of objects to be classified. An amputee generally has a very limited ability to create different patterns of EMG signals, which means a limited number of classes (movements) that we can include in the recognition system. For example, the full mobility of a dexterous prosthesis consists of 35 movements \cite{Cini2019}, while the patient's practical ability is typically limited to controlling no more than 12 movements \cite{Mendez2021}.
This limitation can be counteracted by systematic training of the patient according to a specific protocol based -- for example -- on brain cortex plasticity stimulated by visual (via virtual reality) and sensory feedback procedures \cite{Kurzynski2017} or using the attractive computer game formula \cite{Kristoffersen2021}. It should be emphasised that the factor in question does not disqualify the bionic prosthesis but limits its dexterity and multifunctionality.

More serious consequences can result from incorrect operation of the prosthesis, i.e. making a movement other than intended, caused by misclassification of the patient's intentions. An unacceptable error rate discourages the patient from using the prosthesis, as it becomes of little use (the prosthesis does not follow the user's will) in supporting the user in everyday activities.
Ensuring a high quality of the recognition process is a difficult challenge because many factors originating from the human controller (object source) and the human-prosthesis interface (HPI) significantly distort the information about the patient's intention carried by the recorded EMG signals. 
The following factors have a particularly noticeable adverse effect on the quality of the recognition process \cite{Trajdos2025,Schone2023,Maurya2025}:
\begin{enumerate}
\item \textbf{Non-stationary EMG signals} caused by: electrode shifts, changes of arm postures, changes of electrode/skin impedance (sweating), doffing/donning, muscle fatigue, change in fibre geometry;
\item \textbf{Contamination of EMG biosignals}.  Three main types of signal contaminants can be distinguished:
\begin{enumerate*}[label=(\arabic*)]
    \item  noise (thermal and flicker noise, amplifier saturation, analogue to digital signal clipping, quantisation noise),
    \item interference (power line, ECG, crosstalk),
    \item artifacts (measurement artifacts, baseline wander, motion artifact, poor electrode placement);
\end{enumerate*}

\item \textbf{Low intra-subject repeatability in generating EMG signal patterns} --  changes due to human motor learning/adaptation of the user to the control of the prosthesis. This more generally denotes concept drift, which is related to the change in the model (concept) of the object source (human); 
\item \textbf{Inherent instability of signal}. Caused by the variable frequency of motor unit firing (0-20 \si{\hertz}).
\end{enumerate}

In addition, there are other circumstances that should be mentioned, all of which mean that we are dealing with a difficult recognition problem \cite{Kyranou2018, RodriguezTapia2020}: 
\begin{enumerate}
\item Curse of dimensionality: a large number of primary features extracted from multimodal and multichannel biosignals; 
\item A small learning data set: difficulties in obtaining a sufficient set of training data  due to phantom pain;
\item Non-oracle teacher: possibility of incorrect labeling of learning objects by the amputee;
\item Real time requirement: the control system should produce a prediction of the intended movement with max delay = 200\si{\milli\second}.
\end{enumerate}

Numerous papers addressing the application of various classification models and feature engineering procedures have been published. They are summarised in synthetic review publications~\cite{Parajuli2019,Chen2023}. In this domain, the performance of advanced classification techniques and recognition systems featuring a variety of architectures and functionalities has been evaluated. This encompasses multistage recognition involving the decomposition of the classification process \cite{Kurzynski2016}, recognition through multi-classifier systems with diverse combination strategies \cite{Kurzynski2017,Akbulut2022}, data stream recognition accommodating concept drift \cite{Vidovic2016} and contextual recognition \cite{Trajdos2024}. 

Despite the great activity of the scientific community in the development of methods for biosignal-recognition-based control of the bionic hand, the problem remains open to further research.

In the paper, the authors propose a novel compound classification system that recognises the intention of the patient on the basis of multichannel EMG biosignals and identifies contaminated signals (channels). The system consists of two ensemble classifiers. The first classifier is created by a set of OCC, the purpose of which is to assess the possible contamination of individual biosignals.  The second classifier is an ensemble of KNN models that finally recognises the patient's intentions. For both recognition systems, an original and coherent fuzzy model was developed, which allows the use of a uniform soft decision scheme throughout the recognition process.
This concerns the fuzzy description of the degree of signal contamination with different fuzziness models determined by the OCC and the fuzzy similarity measures used in the KNN system.

The main research questions that are to be answered are as follows:
\begin{enumerate}[label=\textbf{RQ \arabic*}]

  \item \label{itm:rq:ent_feat_space} Does the proposed method perform better than a simple classifier trained on the entire feature space?

  \item  \label{itm:rq:ltr_comp}Does the proposed method perform better than the alternative approaches presented in the literature?

  \item \label{itm:rq:snr_vals} How do the investigated methods perform under different values of the signal-to-noise ratio (SNR)?
  
  \item \label{itm:rq:fmember} How do the proposed method perform under different fuzzy models of target class of one-class classifiers?

\end{enumerate}

The remainder of the article is organised as follows.
Section~\ref{sec:RelWorks} reviews the previous works related to the topic of this paper. Section~\ref{sec:ProposedMethod} contains a comprehensive description of the proposed method.
In Section~\ref{sec:ExperimentalSetup}, which describes the experimental protocol, we present the signalsets, the comparative studies protocol, the extraction methods, the selection of features, and the classifier models.
The results obtained are presented and discussed in Section~\ref{sec:resultsanddisc}. Finally, concluding remarks are presented in Section~\ref{sec:Conclusions}.

\section{Related Works}\label{sec:RelWorks}

The review of the literature related to this paper will be divided into two sections. The first includes works on the methodological aspect based on fuzzy methods in application to the control of a bioprosthetic hand.
The second section is focused on the improvement of the sEMG signal using the ML-based approaches. 

\subsection{Fuzzy methods applied to the control of hand bioprosthesis}\label{sec:RelatedWorks:fuzzy}

In myoelectric control of the prosthesis of the upper limb, two basic approaches are distinguished \cite{Parajuli2019}:
\begin{enumerate*}[label=(\Alph*)]
    \item Non-recognition-based control in which the degrees of freedom (DoF) of the prosthesis are directly controlled by sEMG signals; 
    \item Recognition-based control, where the recognition system first classifies sEMG signals and then, based on the  classification results, the kinematic controller controls DoF  of the prosthesis. 
\end{enumerate*} 
There are numerous publications in the literature that present control and recognition algorithms based on fuzzy set theory for all decision problems that occur in the above-mentioned approaches. Their overview, divided into decision-making tasks, is described in the following subsections.

\subsubsection{Non-recognition-based controller}\label{sec:RelatedWorks:fuzzy:nonrecog_controller}
Fuzzy logic controllers, proposed over 50 years ago, have been used in a wide variety of practical control systems, demonstrating their numerous advantages. They were also used in direct control of an upper limb bioprosthesis and tested under laboratory conditions. The most popular are fuzzy controllers of Mamdani type (MFC) and Takagi-Sugeno type (TSFC). The scheme of both models includes a stage of fuzzification of crisp input data (antecedents), an inference engine based on a set of fuzzy rules of Mamdani/Takagi-Sugeno type usually built in a supervised learning procedure or using expert knowledge, and a stage of defuzzification of the output quantities (consequents) to obtain a crisp decision value~\cite{Nguyen2019}.

In \cite{Duan2015} the authors proposed multivariable MFC for control of multi-finger prosthetic hand.  In order to simplify the controller, all variables share a common fuzzy rule base and identical membership functions. The MFC consists of fuzzifier (singleton method), inference engine (fuzzy logic used to combine fired fuzzy rules of Mamdani type into a mapping from fuzzy input sets to fuzzy output sets), and defuzzifier (the centre-of-sets method). Experiments were conducted in a LabView environment using simulated five-fingers (each finger had three knuckles) hand. Input values were produced by sensors contacting the grasped object. The simulation demonstrated that the control process using mulivariable MFC is effective in terms of time and quality of control.

The hand prothesis control system presented in \cite{Tabakov2016} is based on MFC with parameters of sEMG signals acquired from two amputees (input values) and changes in the speed of a hypothetical motor unit (output values). The experiments carried out show results comparable to those of the classical PID controller. 

In \cite{Precup2018} the authors propose applying TSFC for the control of the hand prosthesis. In the proposed approach, the bionic prosthesis is a multi-input-multi-output (MIMO) non-linear dynamical system, with the inputs represented by the sEMG signals and the outputs
represented by finger angles at various joints. The authors investigated models of finger dynamics and, as a result in consequences of fuzzy rules of Takagi-Sugeno type, linear functions of first and second order were proposed. In antecedents of TS fuzzy rules are linguistic terms of features of EMG signals captured from 8 sensors.  The weighted average defuzzification method was used to obtain the crisp control (output) variables. The simulation results conducted using the best fuzzy models of the process highlight the performance improvement of the control system with fuzzy controllers versus the control
system with linear ones.

The paper \cite{Qishqish2020} presents a comparative analysis of the PID controller,  Mamdani fuzzy controller, and Genetic algorithm fuzzy logic controller (GAFLC), in terms of the response of the system to be controlled. The simulation of hand prosthesis control was performed using a Matlab / Simulink environment. The GAFLC system is obtained from MFC in which arbitrary parameters are optimised by the GA method. The simulation results show that the MFC with the GA optimisation procedure is better than PID and MFC in terms of the overshoot.

\subsubsection{Recognition-based control -- classification system}\label{sec:RelatedWorks:fuzzy:recog_controller}

In the area of fuzzy-based sEMG signal classification, solutions using the Mamdani scheme are dominant. This is due to the fact that TSFS systems are not suitable for the discrete decision problem of classification.

The paper \cite{Chan2000} proposes a fuzzy decision model to classify the single-site sEMG signal for multifunctional prosthesis control. Time segmented features are fed to a fuzzy system for training and classification. Gaussian membership functions and the center of gravity method were employed in the fuzzification and defuzzification procedures. Fuzzy rules of the system were trained using the clustering algorithm (the Basic Isodata) and the back propagation method. 
Experiments were conducted for 4 healthy subjects and recognition problem with 4 classes (flexion, extension, pronation,  and supination). sEMG signal was divided into 6 segments, and each segment was represented by 4 features (MAV, ZC, SSC, and WC). The results (accuracy from 70 to 98 percent) were compared with the 3 layer ANN classifier. 
 
In \cite{Chauvet2001} the authors consider another problem of recognising the sEMG signal, which can also be used in the prosthesis control procedure.
It is a two-class task in which it is necessary to recognise whether the constituent motor unit action potential (MUAP) occurred in the sEMG signal or not. The authors propose a specific iterative algorithm with a classification method using fuzzy logic techniques with the following parameters: three input variables (inter-pulse interval, amplitude, corelation between class and sEMG shape), trapezoid membership functions, 27 fuzzy rules and Zadeh inference engine. 
For simulated signals, the rate of successfully classified MUAP was 88.4 percent. For real sEMG signals obtained from a healthy subject and with a Laplacian sensor located over the \textsl{biceps brachii}, the algorithm correctly identified 21 MUAPs on the 29 MUAPs identified by an expert.

In \cite{Weir2003} the Mamdani fuzzy decision system was used to detect the onset of sEMG and classify user intent in the following sEMG recognition problem: 4 sEMG registration channels, Hudgins features in time/frequency domain, 4 classes (wrist extension, wrist flexion, ulnar deviation, finger flexion). The fuzzy decision system was implemented using the MATLAB environment with triangular/trapezoid membership functions and automatically generated fuzzy rules using the fuzzy clustering technique.
Experiments were conducted on the real sEMG signals coming from two individuals without amputation and two persons with transradial amputations. Fuzzy systems produced results of 91 to 95 percent accuracy for the amputees and 96 to 98 percent, for the intact-limbed subjects, respectively.

In turn, in paper~\cite{Ajiboye2005} the Mamdani fuzzy system was applied to the following recognition task for prosthesis control: 3 channels of registration of sEMG signals, basic signal statistics (mean and standard deviation) as a feature vector, and 4 classes (prosthesis movements). In the fuzzy logic system, triangular membership functions were used, and fuzzy c-means  data clustering was used to automate the construction of a fuzzy rule base. 
Experiments were conducted on 4 amputees -- overall classification rates ranged from 94 to 99 percent of accuracy. The fuzzy algorithm also demonstrated success in real-time classification achieving decision time less than 50 ms.

In \cite{Ahmad2012}  the Mamdani fuzzy system was compared to the ANN classifier. The study used various models of membership functions for the input data and a triangular membership function for the output. The centroid method was used for defuzzification. The results (accuracy over 97 percent) demonstrated the superiority of the fuzzy approach over the ANN method.

Similar investigations of the Mamadani fuzzy system are presented in \cite{Ulkir2017}. In experiments for 2 sEMG signals registered over extensor and flexor muscles 3 features (RMS, WL (wavelet length) and kurtosis) were extracted. In the inference engine, the triangular membership function was used and 6 IF-THEN rules were constructed on the basis of the learning set. The classification accuracy obtained by the system was 93 percent.

The research presented in \cite{Taar2017} includes the design of an anthropometric prosthetic hand using 3D printing, as well as the implementation of a motion class recognition algorithm based on the Mamdani fuzzy classifier. The recognition task used sEMG signals recorded from 4 electrodes, from which 5 (per channel) time-frequency domain features were extracted. In the fuzzy system, the sum of features from one channel was used as an input value with triangular membership functions in the fuzzification procedure and the centroid method at the defuzzification stage. The set of 28 fuzzy rules was created based on information from the states of contraction. Experiments for 6 class problem (finger movements - hand closure/opening, index/middle/ring/pinky thumb touch) on real data coming from one able-bodied person gave results from 100 percent (hand closure/opening) to 78,52 percent of accuracy (middle-thumb touch).

In \cite{Huang2023} the authors proposed an original two-level classifier for the control of the bioprosthetic hand. Grasp posture classification is divided into first-level and second-level fuzzy decision strategies. The first-level classifier is used to determine the grasp posture preliminarily according to the size and roundness of the object from the visual feedback. The second-level classifier integrates visual and sEMG decision results to jointly
determine the final grasp posture. Both classifiers are based on the classical Mamdani system and used the set of 25 fuzzy rules built on the basis of the transfer learning procedure.  For antecedents and consequent at both levels, triangular and trapezoid membership functions were applied. 
Experiments were conducted on real data coming from 3 subjects with upper limb amputation and for 5 hand gestures (relaxation, flat, thumb up, wrist extension, wrist flexion). The results show that the proposed method can improve the grasp posture classification.

Mamdani's decision systems are not the only method with a fuzzy model that has been practically applied in the task of recognising sEMG biosignals for bioprosthesis control. Another approach is neuro-fuzzy systems (NFS). A hybrid NFS system combines a neural network and fuzzy logic in one structure, so it has the potential to demonstrate the advantages of both methods. For this reason, it is readily used in practical problems, including recognition tasks.

In \cite{Karlik2003} the authors propose a fuzzy clustering neural network (FCNN) as a classifier of sEMG signals applied in the task of controlling multifunctional prostheses. This method was experimentally compared with ANN with back-propagation procedure and conic section function neural network.
The sEMG signals were used to classify six upper-limb movements: elbow flexion/extension, wrist pronation/supination, grasp, and resting. The results obtained using the proposed method are the best and amount to 98 percent of accuracy.

The aim of the paper \cite{Khezri2007} was to apply an adaptive neuro-fuzzy inference system (ANFIS) to recognise sEMG patterns in the control of an upper limb prosthesis. Experiments were conducted on real sEMG signals from 4 able-bodied subjects, and classification problem contained 5 time/frequency domain features and 6 hand movements (hand closing/opening, pinch and thumb flexion, wrist flexion/extension). The accuracy of the ANFIS method was equal to 96.7 percent.

In the work \cite{Fariman2015}, similarly to the previous paper, the ANFIS system was also used to classify sEMG signals. In experimental investigations, sEMG signals were acquired from 2 forearm muscles of 4 subjects in the normal limb. These signals were segmented and the features were extracted using a combined time-domain approach. Movements were classified using the ANFIS method and the ANN classifier. Comparative analysis indicates that ANFIS not only displays higher classification accuracy (88.90 percent) than the ANN, but it also improves computation time significantly.

\subsubsection{Recognition-based control -- kinematic controller}\label{sec:RelatedWorks:fuzzy:recog_kincontroller}
In \cite{Barfi2022} the authors investigated the bioprosthesis control system by recognising patient's intentions, in which a fuzzy controller was used at the kinematic control stage. The experimental setup was as follows:
\begin{enumerate*}[label=(\arabic*)]
    \item two-channels sEMG signals were collected from the forearm of 5 able-bodied persons;
    \item  for each channel two features (MAV and VAR) were extracted;
    \item 8 prothesis movements were considered (cylindrical, hook, lateral, point, rest spherical, tripoid, tip);
    \item 3 models were used in the classifier that preceded the kinematic controller (LDA, QDA, kNN, SVM).
\end{enumerate*}
All movements were modelled in a new way through
angels of three joints for each finger, thus the adaptive fuzzy-PI controller was used to control an artificial hand with fifteen degrees of freedom. The results showed that the fuzzy-PI controller performs better than the classical PI controller. 

\subsection{Methods of mitigating contami- nations in the sEMG signal }\label{sec:RelatedWorks:mitigation}

The quality of the surface electromyographic (sEMG) signals is very important in the task of controlling a bionic hand prosthesis. When environmental or physiological
factors (for example, muscle fatigue, skin perspiration) degrade the signal, controlling the device becomes more difficult \cite{Farago2023}. 

One of the most interesting categories of methods, which have been proposed in the literature, employ machine learning techniques to address signal disturbances. 

The approaches in this category can be further divided into several subcategories. The first subgroup is comprised of techniques solely focused on identifying disrupted signals and preventing the prosthesis from performing movements triggered by these erroneous signals~\cite{Ding2022}. Multiclass classifiers must have both contaminated and clean patterns in the training dataset to build the model~\cite{Irfan2023, Jena2024}. Such methods might utilise various classifiers to identify signal disturbances, for example SVM~\cite{Irfan2023, Jena2024} or neural network models~\cite{Machado2020,Jena2024}. However, it requires the prediction of the types of noise pattern using expert knowledge prior to the training phase. As a result, such methods are only capable of identifying specific types of disturbance, leaving other types undetected. To address this limitation, one-class classification can be employed. This machine learning approach is tailored to identify anomalies in cases where the training data consists exclusively of clean signals. For example, the technique introduced by Freaser et al. identifies disturbances in channels through the application of a one-class SVM classifier~\cite{Fraser2014}. The anomalies can also be detected using approaches based on unsupervised learning. The authors of~\cite{Ijaz2018} used self-organising maps and hierarchical clustering to detect outlying signals. The outlying signals were assumed to be the noisy ones. It is also possible to identify contaminated signals and use them to modify the operating scheme of the classification system. Furukawa et al. introduced a multi-classifier system designed to tolerate contamination in a single sEMG channel~\cite{Furukawa2015}. For a system with eight sEMG channels, an ensemble of eight distinct classifiers was developed, where each classifier operates using only seven of the channels. When one of the channels is contaminated, a proper seven-channel classifier is used (one that does not use the contaminated channel). They also cover cases where no disturbances are found, in which a classifier built using all eight sEMG channels is used. This approach has a major drawback since it assumes that only one sEMG channel can be contaminated. To overcome this limitation, a dual-ensemble approach has been proposed that allows contamination of all channels except one~\cite{Trajdos24b}.

 Some researchers suggest retraining the whole classification system after the detection of compromised channels. However, this method is impractical because it potentially requires numerous retraining sessions~\cite{Reynolds2021}. These systems are used mainly in scenarios where a sensor defect occurs, leading to infrequent retraining processes. Researchers developing this kind of algorithm aim to accurately detect faulty sensors and achieve a short retraining time~\cite{Reynolds2021}. Alternatively, a classifier system that does not require direct detection of noisy channels could be proposed. This approach involves employing a robust classification ensemble; for example, a classifier based on error-correcting output codes is proposed in~\cite{Sarabia2023}. The authors of \cite{Favieiro2025} suggest using a modified version of the random forest classifier, which is inherently robust to noise. To make the classifier more resistant to noise, they propose a method based on paraconsistent logic.
 
 Another approach involves incorporating data with simulated noise to develop a resilient classifier system~\cite{Lin2023}. Transfer learning techniques are also possible to adapt the classifier to new conditions, such as noise contamination~\cite{Ameri2020,Tosin2025}.

It is also possible to detect contaminated signals and then try to remove contamination using signal processing techniques. The authors of~\cite{AitYous2024} propose a two-stage fuzzy inference system to identify and eliminate noise from sEMG signals. In the first stage, the system detects whether the signal is contaminated. If contamination is detected, the signal is decomposed using one of several decomposition techniques (DWT, SWT, EMD, VMD). In the second stage, another fuzzy inference system determines which of the decomposed elements contain noise. The final clean signal is reconstructed by combining only the elements identified as clean.

In this work, we introduce a multi-classifier framework that utilises a dynamic ensemble selection mechanism driven by the results from a one-class classifier ensemble designed to detect noisy signal channels. This approach is an extension of the technique described in~\cite{Furukawa2015,Trajdos24b}.

\subsection{Summary}\label{sec:RelatedWorks:summary}
Two conclusions can be drawn from the review of the literature presented:
\begin{enumerate}
\item Fuzzy logic-based models that have been applied to decision-making problems relevant to bioprosthetic hand control schemes are represented by fuzzy decision systems with a typical Mamdani or Takagi-Sugeno architecture and fuzzy-neural approach. The decision-making model based on fuzzy relations proposed in this paper is not among them;
\item The software solutions for mitigating sEMG signal contamination proposed in the literature, based on broadly understood artificial intelligence methods, do not include ideas using fuzzy logic in decision-making processes.
\end{enumerate}

To the best of the authors' knowledge, in \cite{Trajdos2025} a constructive fuzzy model of a compound recognition system composed of many cooperating classifiers applied to bioprosthesis control was presented for the first time. This paper is a creative extension of the results presented therein with a new coherent fuzzy model for a different recognition scheme, classifier models, and type of recognised biosignals.

\section{Proposed Method}\label{sec:ProposedMethod}

Let us consider a system for the control of a bionic upper limb prosthesis based on the multichannel EMG signals recognition scheme. Let
\begin{equation}   \label{1}
\mathcal{C}=\{C_1, C_2, \ldots,C_L\} 
\end{equation}  
denote the set of EMG signals recorded from $L$ sensors (channels) located on the patient's forearm stump. 

For each EMG signal from the set \eqref{1} feature extraction and feature dimensionality reduction  procedures are applied that collectively transform the recorded signal $C_l$ into a feature vector $x_l$ in the feature space $\mathcal{X}_l$ ($l=1,2,\ldots, L$).
Let then
\begin{equation}   \label{2}
\mathcal{M}=\{1,2, \ldots , M\}
\end{equation}
denotes a set of class numbers (labels).
The number of classes $M$ for each user may be different, as it is related to the size of the PME repertoire and the user's ability to activate the stump muscles.

The proposed compound recognition system includes two multiclassifiers whose base classifiers are trained using a supervised learning scheme.  
For the recognition of signal $C_l$ ($l=1,2,\ldots,L$), let
\begin{equation}   \label{3}
X_l=\{x_{l,1}, x_{l,2}, \ldots , x_{l,N}\}, \ \ x_{l,n} \in \mathcal{X}_l 
\end{equation}
be the learning set, i.e. the set of labeled samples. Let $X_l^{(j)} \subset X_l$ denotes learning samples from $j$th class ($j \in \mathcal{M}$). We assume that the learning samples \eqref{3} are recorded without contamination under laboratory conditions. 

The details of both multiclassifier systems are provided in the following subsections.

\subsection{Ensemble of one-class classifiers}\label{sec:ProposedMethod:oneclass}

The first multiclassifier system consists of an ensemble of one-class classifiers:
\begin{equation}    \label{4}
\Phi = \{ \phi_1, \phi_2, \ldots , \phi_L\},
\end{equation}
where base classifier $\phi_l$ operates in the space $\mathcal{X}_l$ of features extracted from the signal $C_l$ ($l=1,2, \ldots,L$). 
During the classification phase, the base classifier $\phi_l$ assigns an unknown object described by features $x_l$ to the target class $T_l \subset \mathcal{X}_l$ or considers the object an outlier. 

In the considered task, the target class consists of biosignals recorded without contamination.   This means that the classifier $\phi_l$ decides whether the biosignal $C_l$  was recorded without or with contamination.
 In other words, the OCC system returns information on the contamination / purity of individual biosignals \eqref{1}. 

According to the OCC training rule, the classifier $\phi_l$ is trained only using objects from the target class. For this purpose, we will use the training objects \eqref{3} without their labels from the set \eqref{2}.

Unlike the classic (crisp) decision scheme of OCC, we will adopt a soft scheme based on a fuzzy model introduced for the task of recognising noise-contaminated signals in~\cite{Trajdos2025}. In this model, it is assumed that the target class $T_l$ is a fuzzy set in the universe $\mathcal{X}_l$. Consequently,

\begin{equation} \label{5}
\begin{split}
T_l = \{ (x_l, r_l(x_l)) \; | & \; x_l \in \mathcal{X}_l, r_l \in [0, 1] \}\\
 &, l = 1, 2, \dots, L
\end{split}
\end{equation}
where $r_l: \mathcal{X}_l \rightarrow [0, 1]$ is a membership function of $T_l$. 
If we assume
\begin{equation}   \label{6}
\phi_l(x_l) = r_l(x_l),
\end{equation}
then defining OCC $\phi_l$ comes down to determining the membership function of the target class set $T_l$. 

\begin{figure*}[htb]
    \centering
        \includegraphics[width=0.7\textwidth, height=.4\textheight, keepaspectratio]{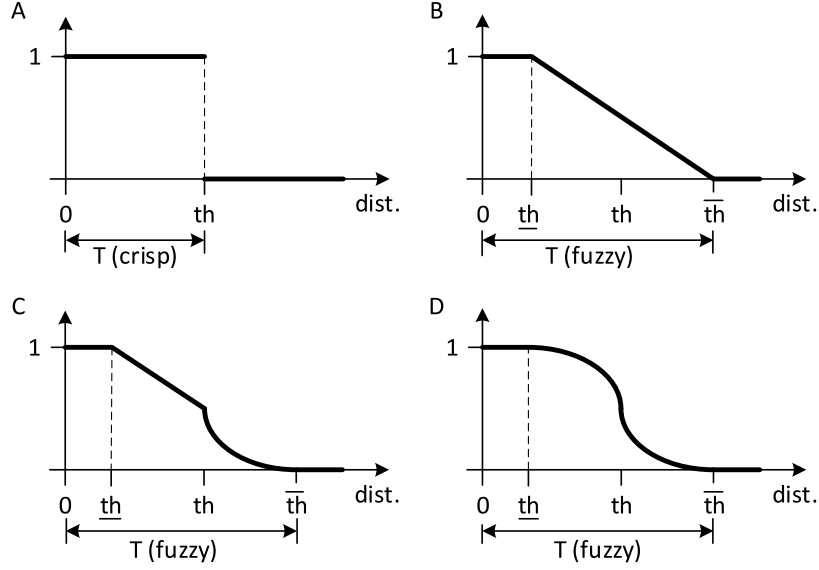}
    \caption{Examples of membership function $r(x)$ of set $T$ --  crisp set: (A) threshold function; fuzzy set: (B) linear function; (C) spline function (linear and square); (D) spline function (square and square).}
    \label{figs:Fig1}
\end{figure*}

In the classical crisp approach, the operation of an OCC is based on determining a surface in the feature space that separates the target class set from the area of outliers \cite{Krawczyk2014}. 
The recognition result is uniquely determined by the position of the object relative to the separating surface.
In this case, we are dealing with a crisp decision scheme in which the function $r(x)$ in \eqref{6} defining the basic one-class classifier is a characteristic function (indicator) of the set $T$, which is schematically illustrated in Figure~\ref{figs:Fig1}A. The variable $x$ in the drawings has the meaning of distance (dist.) from the conventional center of the target class, and the location of the separation point is symbolised by the threshold ($th$). The following Figures~\ref{figs:Fig1} (B, C and D) illustrate the cases of the fuzzy model for different forms of the membership function $r(x)$. Now for the fuzzy set $T$ we deal with the interval $[\underline{th}, \overline{th}]$, in which the membership function decreases from 1 to 0. The presented forms of the membership function correspond to the functions $r(x)$ adopted in experimental studies.

The interpretation of the fuzzy set $T_l \subset \mathcal{X}_l$ is the same as for the crisp scheme: it is a set of objects (feature vectors $x_l$) extracted from signals $C_l$ recorded without contamination. However, now we have additional information on the degree of purity (contamination) of signal $C_l$, which is the value of the membership function at the point $x_l$. This information will be used in the next ensemble system.

\subsection{Ensemble of fuzzy KNN classifiers  with a dynamic fusion procedure}\label{sec:ProposedMethod:multiclassifier}

The second classification system consists of $L$ KNN classifiers with the original fuzzy model. 
The choice of such a classifier is deliberate and justified by the fact that the KNN method belongs to the category of lazy classifiers. The KNN model is built on the fly, locally, for each recognised object, demonstrating high flexibility and easy adaptation to new data. Since in the considered task there is a dynamic selection of features (depending on the identified contamination level), the properties of lazy learning are particularly useful here, significantly accelerating the training process. A similar effect was obtained using the Naive Bayes classifier, in which the multiplicative form of the model allowed a one-shot learning mode \cite{Trajdos2025a}.

Lets now define a binary fuzzy relation between the feature space $\mathcal{X}_l$ and the learning set $X_l$ as a fuzzy subset of Cartesian product $\mathcal{X}_l \times X_l$, namely:
\begin{equation} \label{7}
\begin{split}
S_l = \{ (x_l, x_{l,n}),\; sim(x_l, x_{l,n}) \; | & \; x_l \in \mathcal{X}_l, \; x_{l,n} \in X_l, \\
                                                  & \; sim(\cdot , \cdot) \in [0, 1] \}
\end{split}
\end{equation}
The relation $S_l$ describes the connection:
 ''$x_l$ is similar to  $x_{l,n}$'' (or vice versa) and $sim(\cdot , \cdot)$ is a normalised similarity measure in $\mathcal{X}_l$.

Similarity measures are closely related to distance measures. In \cite{Levy2024} we can find a survey of more than fifty related distance and similarity measures, categorised by measure families. 

The similarity measure used in the experiments is based on a Gaussian kernel \cite{Metcalf2016}:
\begin{equation}\label{eq:simfun:def}
    \simfun(x_l, x_{l,n})  = \exp\left(-\frac{\eucdist{x_l}{x_{l,n}}^2}{2\sigma^2}\right),
\end{equation}
that is a nonlinear function of the Euclidean distance $\eucdist{x_l}{x_{l,n}}$ decreasing with distance and ranges between zero and one. $\sigma$ is the standard deviation of all distances in the training set. 
The Gaussian kernel provides a relatively simple way to measure the similarity between data points in a high-dimensional space, and therefore it is often used in practical data mining problems.

Now create a new fuzzy set as an intersection of sets $T_l(x_l)$ and $S_l(x_l, x_{l,n})$:

\begin{equation}  \label{8}
U_l(x_l, x_{l,n}) = T_l(x_l) \cap S_l(x_l, x_{l,n}).
\end{equation}

The set \eqref{8} is still interpreted as a fuzzy relation '' $x_l$ is similar to $x_{l,n}$'', but now we additionally take into account the uncertainty of this relation due to the noisy measurement of the point $x_l$.
This is visible in the membership function of the set $U_l$, which we take as the algebraic t-norm:
\begin{equation}   \label{9}
u_l(x_l, x_{l,n}) = r_l(x_l) \cdot \simfun(x_l, x_{l,n}).
\end{equation}
The factor $r_l(x_l)$ corrects the degree of similarity between $x_l$ and $x_{l,n}$ in proportion to the degree of uncertainty of the true position of point $x_l$ in the feature space $\mathcal{X}_l$. 
The membership function \eqref{9} can then be called a corrected similarity measure that is not distorted by the noisy measurement of $x_l$. 

Let next for fixed $x_l=x_l^0$: 
\begin{equation}   \label{10}
U_l[\alpha_K, x_l^0](x_{l,n}, u_l(x_l^0, x_{l,n}) )=
\end{equation}
\begin{equation}   \nonumber
=\{x_{l,n}, u_l(x_l^0, x_{l,n})| x_{l,n} \in X_l,  u_l(x_l^0, x_{l,n}) > \alpha_K\}
\end{equation}
denotes the $\alpha_K$-cut fuzzy set  which  contains $K$ learning objects from $X_l$. The $\alpha_K$-level is between $K$-th and ($K+1$)-th the most similar (in the sense of measure $u_l$) learning patterns to $x_l^0$.

Finally, we define a partition of the set \eqref{10} as a collection of disjoint fuzzy subsets:
\begin{equation} \label{11}
\begin{split}
U_l[\alpha_k, x_l^0](x_{l,n}, u_l(x_l^0, x_{l,n})) 
&= \bigcup_{j=1}^M V_l^{(j)}[\alpha_k, x_l^0] \\
&\quad (x_{l,n}, u_l(x_l^0, x_{l,n}))
\end{split}
\end{equation}
where for $j \in \mathcal{M}$:
\begin{equation}   \label{12}
V_l^{(j)}[\alpha_k, x_l^0](x_{l,n},u_l(x_l^0, x_{l,n}))= 
\end{equation}
\begin{equation}   \nonumber
=\{x_{l,n}, u_l(x_l^0, x_{l,n}) | x_{l,n} \in X_l^{(j)},  u_l(x_l^0, x_{l,n}) > \alpha_K\}  
\end{equation}
are defined by means of the equivalence relation of learning objects belonging  to the same class \cite{Hajnal2009}.

By summing the cardinality of the sets \eqref{12} over all EMG biosignal channels and normalizing the result to be within the range of $ [0, 1]$, we obtain the classification functions of the ensemble of KNN classifiers for the object $x^0=(x_1^0, x_2^0, \ldots, x_L^0)$ to be classifed:
\begin{equation}  \label{13}
d_j(x^0)= \frac{\frac{1}{L}\sum_{l=1}^L | V_l^{(j)}| }{\sum_{j=1}^M {\frac{1}{L}\sum_{l=1}^L  |V_l^{(j)}|} }, \ \  j=1,2,\ldots,M.
\end{equation}

The operation of the proposed compound recognition system with a consistently applied fuzzy model is fully intuitive and can be summarised in the following sentence: Assign an object to the class in which the sum of similarity measure corrected by the contamination level of the measurement over all recording channels is the largest among the $K$ most similar training objects.

\section{Experimental Setup}\label{sec:ExperimentalSetup}

The experimental study is conducted to answer the research questions posed in Section~\ref{sec:Introduction}.
To do so, we performed three main experiments. The first is to compare the proposed fuzzy KNN algorithm with a similar attribute weighting approach using the Naive Bayes Clasifier~\cite{Trajdos2025a}. The second assesses the performance of the proposed method with different membership functions of the fuzzy set $T_l$. The third is to compare the proposed method with reference methods taken from the literature.

\subsection{Common Setup}\label{sec:ExperimentalSetup:Common}

The signals used in these experiments come from the Web repository \footnote{\url{https://www.rami-khushaba.com/}}. The dataset contain data from nine different subjects with amputation. The more precies information for each amputee and the data set details are presented in~\cite{AlTimemy2016}. 
To standardize the signals from various subjects, we utilized sEMG signals from the initial 8 channels, focusing on those related to a low force level. For distinguishing individual objects within specific classes, a non-overlapping segmentation approach was employed with a segment duration of 500 ms. Consequently, this produced between 70 and 310 objects per class.

The methods investigated in the experimental study are compared using the following quality criteria:
\begin{itemize}
    \item Balanced Accuracy criterion (BAC);
    \item Cohen's Kappa criterion (Kappa);
    \item Micro-averaged $F_1$ criterion (F1).
\end{itemize}

As an outlier detector, we use a one-class SVM classifier. This classifier proved to be a useful tool in the task of detecting contaminated EMG signals~\cite{Fraser2014}. The $\nu$ parameter of this classifier is tuned using the following procedure. Using a four-fold cross-validation procedure, the training and validation sets are extracted. The validation (testing) part is then augmented with artificial examples marked as noise. Artificial examples are generated using a uniform distribution in the classifier-specific input space $\mathcal{X}_l$. The following values of $\nu$ are considered $\{ 0.1, 0.2, \ldots, 1.0\}$. The classification quality is assessed using the balanced accuracy criterion. When the best value of $\nu$ is found, the final one-class classifier is trained using the entire training set. We use the base classifiers implemented in the scikit-learn package. Unless otherwise specified, the classifier parameters are set to their default values.
To simulate real-world EMG signal contaminations, the following noise generation techniques are used~\cite{Boyer2023}:
\begin{itemize}

\item Simulated power grid noise -- a sinusoidal signal with frequency [48 to 52]\si{\hertz}.

\item Signal attenuation that simulates the sensor losing contact with the skin.

\item Gaussian noise. Simulates the general noises that may appear in the signal acquisition circuit.

\item Simulation of non-linear amplifier characteristics for signals of high amplitude. This is done by non-linear clipping the peaks of the signal.

\item Baseline wandering. This is simulated by injecting low-frequency sinusoidal noise (0.5 to 1.5\si{\hertz}).

\end{itemize}
In experimental studies, we consider the following SNR levels $\{ 0,1,2, \ldots, 6, 10, 12 \}$. The amount of noise injected into the signal is controlled by the SNR parameter.

The $\sigma$ parameter of the Gaussian kernel \eqref{eq:simfun:def} is set to the standard deviation of pairwise distances between the training objects in the feature space $\mathcal{X}_l$. 

The number of nearest neighbours $K$ is tuned using a four-fold cross-validation procedure. The following values of $K$ are considered: $\{1,3,5, \dots, 23\}$. The value of $K$ that maximises the balanced accuracy is selected. The final classifier is trained using the entire training set.

The training and testing sets are obtained by a ten-fold cross-validation repeated 4 times. The testing dataset is then randomly contaminated with one of the above-mentioned noise types with a selected SNR level. The number of contaminated channels is also randomly selected from the following set of values $\{1, 2, \ldots, 7 \}$.

Feature vectors were created from raw signals using the discrete wavelet transform technique. The \textit{db6} wavelet and three levels of decomposition were used. Two functions were calculated for the transformation coefficients \cite{MendesJunior2020}: mean absolute value (MAV) and slope sign change (SSC).

Following the recommendations of~\cite{garcia2008extension}, the statistical significance of the results obtained was evaluated using the pairwise Wilcoxon signed rank test. To control family-wise errors (FWER), the Holm procedure was used. For all tests, the significance level was set to $\alpha=0.05$. For some analysis, the average rank approach is also used.

The experimental code is provided in~\footnote{\url{https://github.com/ptrajdos/FKNN2025.git}}. For the experimental setup aspects not covered in this section, we refer the reader to this repository.

\subsection{Comparing KNN with weighted attributes and Naive Bayes classifier}\label{sec:ExperimentalSetup:KNNvsNB}

In this experiment, we compare the following approaches:
\begin{enumerate}
    \item \textbf{B} -- base classifier trained using all attribute related weights set to 1.0. This represents the baseline method without noise-related attribute weighting.
    \item \textbf{AW} -- attribute weighted approach with attribute weights set to the values obtained using the one-class classifier.
    \item \textbf{AWc} -- the crisp attribute selection method based on the one-class classifier. In this case, the attribute weights are set to 1.0 for the attributes that are not contaminated and to 0.0 for the contaminated attributes.
\end{enumerate}

Each of the above-mentioned methods is used with the following base classifiers:
\begin{itemize}
\item \textbf{KNN}: the KNN classifier with attribute weighting incorporated into the Euclidean distance measure. 
\item \textbf{GNB}: the Naive Bayes classifier with the Gaussian probability estimator.
\item \textbf{NBM}: the Naive Bayes classifier with the mixture of Gaussian probability estimators. The parameter tuning procedure for this method is described in~\cite{Trajdos2025a}.
\end{itemize}

\subsection{Comparing the proposed method with different membership functions of the fuzzy set $T_l$}\label{sec:ExperimentalSetup:MembershipFunctions}

In this experiment, we compare the performance of the proposed method with different membership functions of the fuzzy set $T_l$ defined in \eqref{5}. The following membership functions are considered to be used with the fuzzy KNN classifier:
\begin{itemize}
    \item \textbf{cr}: crisp/step membership function.
    \item \textbf{cr0}: membership function always set to 1.0.
    \item \textbf{nt}: linear membership function.
    \item \textbf{lp}: spline membership function (linear and parabolic).
    \item \textbf{sm}: smoothstep membership function~\cite{hazimeh2020tree}.
    \item \textbf{ss}: generalised logistic membership function~\cite{Rzdkowski2020}.
\end{itemize}

\subsection{Comparing the proposed method with reference methods taken from the literature}\label{sec:ExperimentalSetup:ReferenceMethods}
In this experiment, the following methods are compared:
\begin{itemize}
    \item \textbf{B} -- the base classifier trained using all attributes without noise-related attribute weighting.
    \item \textbf{DO} -- the method described in~\cite{Trajdos24b} with the parameter $K$ set to 1.
    \item \textbf{DOa} -- the method described in~\cite{Trajdos2025} with the parameter $K$ set to 1.
    \item \textbf{AW} -- the method described in~\cite{Trajdos2025a}.
    \item \textbf{AWc} -- the method described in~\cite{Trajdos2025a} but with a crisp attribute selection.
    \item \textbf{FKNN} -- the proposed method with the fuzzy KNN classifier.
    \item \textbf{FKNNc} -- the proposed method with the fuzzy KNN classifier but with the crisp membership function of the fuzzy set $T_l$.
\end{itemize}
\section{Results and discussion}\label{sec:resultsanddisc}

The results of the conducted experiments are presented in the following subsections. The results are presented in the form of average rank plots and tables. The plots presented in Figures~\ref{figs:Ex1_bac} -- \ref{figs:Ex3_f1}, show the average ranks obtained for each qiality criterion for each each method and for a given SNR level. Tables \ref{table:Ex1_bac_stats} -- \ref{table:Ex3_f1_stats} present the average ranks and the results of statistical tests performed to compare the methods. Each row of a table is related to different SNR values. Each column is related to a different method. The rows of a table contain SNR-value-specific valueos of average ranks. For each row, the subscript contains indices of methods (numbers of columns) that the given method is significantly better. If a row contains a dash (--), it means that the method is not significantly better than any other method.

\subsection{Comparing KNN with weighted attributes and Naive Bayes classifier}\label{sec:resultsanddisc:KNNvsNB}

\begin{figure}[htb]
    \centering
        \includegraphics[width=0.99\columnwidth, height=.3\textheight, keepaspectratio]{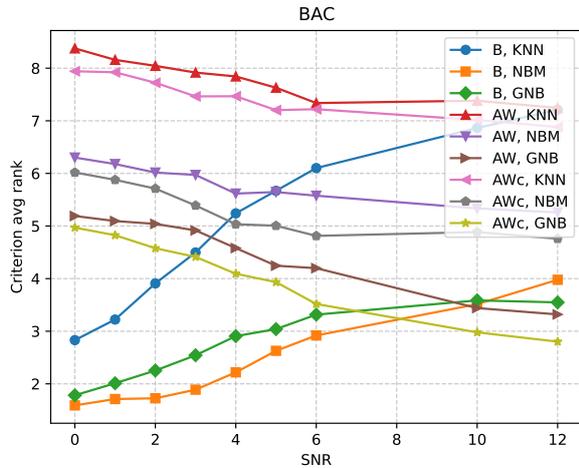}
    \caption{Comparison of the attribute weighting methods. Average ranks plots for Balanced Accuracy criterion.}
    \label{figs:Ex1_bac}
\end{figure}

\begin{figure}[htb]
    \centering
        \includegraphics[width=0.99\columnwidth, height=.3\textheight, keepaspectratio]{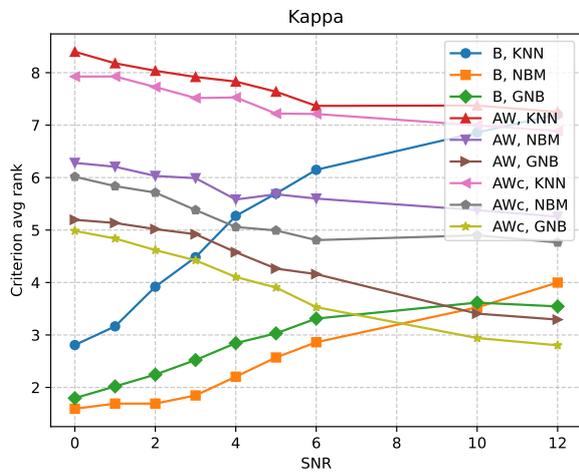}
    \caption{Comparison of the attribute weighting methods. Average ranks plots for Kappa criterion.}
    \label{figs:Ex1_kappa}
\end{figure}

\begin{figure}[htb]
    \centering
        \includegraphics[width=0.99\columnwidth, height=.3 \textheight, keepaspectratio]{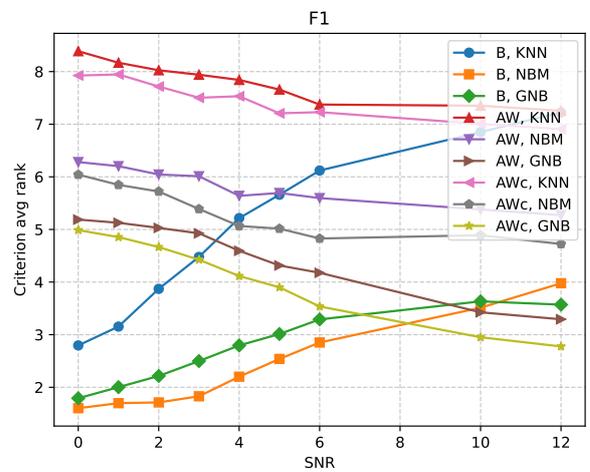}
    \caption{Comparison of the attribute weighting methods. Average ranks plots for $F_1$ criterion.}
    \label{figs:Ex1_f1}
\end{figure}

Let us begin by comparing the base version of the KNN and Naive Bayes classifiers. The results related to this comparison are given in Figures~\ref{figs:Ex1_bac} -- \ref{figs:Ex1_f1} and Tables~\ref{table:Ex1_bac_stats} -- \ref{table:Ex1_f1_stats}. Since all the results for all the quality criteria presented in this section are very consistent, the main trends are discussed together.

In the mentioned case, the results clearly show that for all SNR values, the KNN classifier performs better than both versions of the Naive Bayes classifier. When the SNR value increases, the performance of the KNN classifier improves and the differences between average ranks becomes larger in favour of the KNN algorithm. What is also important, the performance of all base versions of the classifiers increases with the SNR value. This is expected as the amount of noise in the signal decreases with increasing SNR value. However, the KNN classifier benefits more than the Naive Bayes-based approaches. The second observation is that the GNB classifier outperforms the NBM classifier for all SNR values except $\mathrm{SNR} \in \{10,12 \}$. It means that the classifier based on the simpler Gaussian estimator is more robust to noise than the naive Bayes classifier based on the mixture of Gaussians.

The next step is to compare classifiers with weighted/selected attributes. In this case, the results clearly show that KNN with soft attribute weighting is better than KNN with crisp attribute weighting. The results also show that both KNN variants are significantly better than any NB-based algorithm for soft and crisp attribute weighting and for all the SNR values considered. For the NB-based algorithms, the situation is the same. That is, soft attribute weighting gives significantly better results than crisp-weighting (selection) scheme. This observation holds for all SNR values. This observation shows even features formed using contaminated signels carry some useful information that can be useful during the classification process. Consequently, instead of simple removal, features can be used with the appropriate weights to improve classification quality.  The next important observation is that the application of the weighting/selection scheme changes the outcome related to the NB-based classifiers compared to the results without dynamic weighting/selection. We can see that when weighting/selection is applied, then NB-based methods using mixture of Gaussian estimators perform better than the NB-based classifiers using simple Gaussian estimators. It shows that estimators with more parameters perform better when the impact of noisy feratures is reduced.

We should also compare the base classifier versions with variants that perform attribute weighting or selection. The results clearly show that the impact of attribute weighting/selection is the greatest when the amount of signal contamination is the greatest (low SNR values). When the amount of noise is lower, then the differences between base methods and noise-aware methods become smaller in terms of the average ranks. For high SNR values, there are not that many significant differences between base methods and methods that employ attribute weighting selection. For high SNR values, the base and weighted algorithm variants are generally comparable. Only the Naive Bayes classifier with estimators using mixture of Gaussians, the methods that employ weighting/selection are always significantly better than base version. This is a very important observation, as it suggests that the attribute weighting/selection procedure is beneficial even if the signals are not contaminated. It confirms the observations made in~\cite{Trajdos2025,Trajdos2025a}. That is, elimination or attenuation of the influence of contaminated attributes during the classification process improves the performance of the classifier. This is true even if the base classifiers have some robustness to noise and are able to deal with contaminated attributes.
 
{
\setlength\tabcolsep{2.0pt}%
\begin{sidewaystable}[htb]
\centering
\scriptsize
\caption{Comparison of the attribute weighting methods for BAC criterion. Average ranks and statistical tests.\label{table:Ex1_bac_stats}}
\begin{tabular}{l|lllllllll}
 & \multicolumn{9}{c}{BAC} \\

 & {\scriptsize B, KNN (1)} & {\scriptsize B, NBM (2)} & {\scriptsize B, GNB (3)} & {\scriptsize AW, KNN (4)} & {\scriptsize AW, NBM (5)} & {\scriptsize AW, GNB (6)} & {\scriptsize AWc, KNN (7)} & {\scriptsize AWc, NBM (8)} & {\scriptsize AWc, GNB (9)} \\
 \toprule

SNR:12 & 7.210 & 3.978 & 3.549 & 7.244 & 5.258 & 3.319 & 6.882 & 4.758 & 2.801 \\
 & {\tiny 2,3,5,6,7,8,9} & {\tiny 3,6,9} & {\tiny 6,9} & {\tiny 2,3,5,6,7,8,9} & {\tiny 2,3,6,8,9} & {\tiny 9} & {\tiny 2,3,5,6,8,9} & {\tiny 2,3,6,9} & -- \\
SNR:10 & 6.864 & 3.514 & 3.586 & 7.382 & 5.335 & 3.440 & 7.015 & 4.885 & 2.979 \\
 & {\tiny 2,3,5,6,8,9} & {\tiny 9} & {\tiny 9} & {\tiny 1,2,3,5,6,7,8,9} & {\tiny 2,3,6,8,9} & {\tiny 9} & {\tiny 2,3,5,6,8,9} & {\tiny 2,3,6,9} & -- \\
SNR:6 & 6.101 & 2.919 & 3.315 & 7.338 & 5.576 & 4.199 & 7.221 & 4.812 & 3.518 \\
 & {\tiny 2,3,5,6,8,9} & -- & {\tiny 2} & {\tiny 1,2,3,5,6,7,8,9} & {\tiny 2,3,6,8,9} & {\tiny 2,3,9} & {\tiny 1,2,3,5,6,8,9} & {\tiny 2,3,6,9} & {\tiny 2} \\
SNR:5 & 5.669 & 2.628 & 3.040 & 7.629 & 5.646 & 4.246 & 7.203 & 5.006 & 3.933 \\
 & {\tiny 2,3,5,6,8,9} & -- & {\tiny 2} & {\tiny 1,2,3,5,6,7,8,9} & {\tiny 2,3,6,8,9} & {\tiny 2,3,9} & {\tiny 1,2,3,5,6,8,9} & {\tiny 2,3,6,9} & {\tiny 2,3} \\
SNR:4 & 5.240 & 2.218 & 2.907 & 7.843 & 5.617 & 4.581 & 7.467 & 5.035 & 4.093 \\
 & {\tiny 2,3,6,8,9} & -- & {\tiny 2} & {\tiny 1,2,3,5,6,7,8,9} & {\tiny 2,3,6,8,9} & {\tiny 2,3,9} & {\tiny 1,2,3,5,6,8,9} & {\tiny 2,3,6,9} & {\tiny 2,3} \\
SNR:3 & 4.497 & 1.889 & 2.543 & 7.917 & 5.971 & 4.917 & 7.464 & 5.390 & 4.412 \\
 & {\tiny 2,3} & -- & {\tiny 2} & {\tiny 1,2,3,5,6,7,8,9} & {\tiny 1,2,3,6,8,9} & {\tiny 2,3,9} & {\tiny 1,2,3,5,6,8,9} & {\tiny 1,2,3,6,9} & {\tiny 2,3} \\
SNR:2 & 3.908 & 1.725 & 2.253 & 8.044 & 6.017 & 5.040 & 7.724 & 5.711 & 4.578 \\
 & {\tiny 2,3} & -- & {\tiny 2} & {\tiny 1,2,3,5,6,7,8,9} & {\tiny 1,2,3,6,8,9} & {\tiny 1,2,3,9} & {\tiny 1,2,3,5,6,8,9} & {\tiny 1,2,3,6,9} & {\tiny 1,2,3} \\
SNR:1 & 3.222 & 1.711 & 2.011 & 8.158 & 6.178 & 5.094 & 7.922 & 5.878 & 4.825 \\
 & {\tiny 2,3} & -- & {\tiny 2} & {\tiny 1,2,3,5,6,7,8,9} & {\tiny 1,2,3,6,8,9} & {\tiny 1,2,3,9} & {\tiny 1,2,3,5,6,8,9} & {\tiny 1,2,3,6,9} & {\tiny 1,2,3} \\
SNR:0 & 2.831 & 1.590 & 1.782 & 8.376 & 6.301 & 5.190 & 7.942 & 6.018 & 4.969 \\
 & {\tiny 2,3} & -- & {\tiny 2} & {\tiny 1,2,3,5,6,7,8,9} & {\tiny 1,2,3,6,8,9} & {\tiny 1,2,3,9} & {\tiny 1,2,3,5,6,8,9} & {\tiny 1,2,3,6,9} & {\tiny 1,2,3} \\
\end{tabular}
\end{sidewaystable}
}

{
\setlength\tabcolsep{2.0pt}%
\begin{sidewaystable}[htb]
\centering
\scriptsize
\caption{Comparison of the attribute weighting methods for Kappa criterion. Average ranks and statistical tests.\label{table:Ex1_kappa_stats}}
\begin{tabular}{l|lllllllll}
 & \multicolumn{9}{c}{Kappa} \\

 & {\scriptsize B, KNN (1)} & {\scriptsize B, NBM (2)} & {\scriptsize B, GNB (3)} & {\scriptsize AW, KNN (4)} & {\scriptsize AW, NBM (5)} & {\scriptsize AW, GNB (6)} & {\scriptsize AWc, KNN (7)} & {\scriptsize AWc, NBM (8)} & {\scriptsize AWc, GNB (9)} \\
 \toprule

SNR:12 & 7.203 & 3.999 & 3.544 & 7.251 & 5.256 & 3.293 & 6.888 & 4.764 & 2.803 \\
 & {\tiny 2,3,5,6,7,8,9} & {\tiny 3,6,9} & {\tiny 6,9} & {\tiny 2,3,5,6,7,8,9} & {\tiny 2,3,6,8,9} & {\tiny 9} & {\tiny 2,3,5,6,8,9} & {\tiny 2,3,6,9} & -- \\
SNR:10 & 6.860 & 3.526 & 3.617 & 7.374 & 5.379 & 3.407 & 6.996 & 4.900 & 2.942 \\
 & {\tiny 2,3,5,6,8,9} & {\tiny 9} & {\tiny 9} & {\tiny 1,2,3,5,6,7,8,9} & {\tiny 2,3,6,8,9} & {\tiny 9} & {\tiny 2,3,5,6,8,9} & {\tiny 2,3,6,9} & -- \\
SNR:6 & 6.149 & 2.861 & 3.312 & 7.367 & 5.600 & 4.160 & 7.213 & 4.808 & 3.531 \\
 & {\tiny 2,3,5,6,8,9} & -- & {\tiny 2} & {\tiny 1,2,3,5,6,7,8,9} & {\tiny 2,3,6,8,9} & {\tiny 2,3,9} & {\tiny 1,2,3,5,6,8,9} & {\tiny 2,3,6,9} & {\tiny 2,3} \\
SNR:5 & 5.696 & 2.571 & 3.031 & 7.638 & 5.682 & 4.265 & 7.221 & 4.993 & 3.904 \\
 & {\tiny 2,3,5,6,8,9} & -- & {\tiny 2} & {\tiny 1,2,3,5,6,7,8,9} & {\tiny 2,3,6,8,9} & {\tiny 2,3,9} & {\tiny 1,2,3,5,6,8,9} & {\tiny 2,3,6,9} & {\tiny 2,3} \\
SNR:4 & 5.269 & 2.204 & 2.844 & 7.831 & 5.582 & 4.578 & 7.528 & 5.058 & 4.106 \\
 & {\tiny 2,3,6,8,9} & -- & {\tiny 2} & {\tiny 1,2,3,5,6,7,8,9} & {\tiny 2,3,6,8,9} & {\tiny 2,3,9} & {\tiny 1,2,3,5,6,8,9} & {\tiny 2,3,6,9} & {\tiny 2,3} \\
SNR:3 & 4.482 & 1.847 & 2.522 & 7.918 & 5.989 & 4.921 & 7.517 & 5.382 & 4.422 \\
 & {\tiny 2,3} & -- & {\tiny 2} & {\tiny 1,2,3,5,6,7,8,9} & {\tiny 1,2,3,6,8,9} & {\tiny 2,3,9} & {\tiny 1,2,3,5,6,8,9} & {\tiny 1,2,3,6,9} & {\tiny 2,3} \\
SNR:2 & 3.919 & 1.692 & 2.244 & 8.036 & 6.032 & 5.018 & 7.726 & 5.714 & 4.618 \\
 & {\tiny 2,3} & -- & {\tiny 2} & {\tiny 1,2,3,5,6,7,8,9} & {\tiny 1,2,3,6,8,9} & {\tiny 1,2,3,9} & {\tiny 1,2,3,5,6,8,9} & {\tiny 1,2,3,6,9} & {\tiny 1,2,3} \\
SNR:1 & 3.164 & 1.692 & 2.019 & 8.179 & 6.207 & 5.135 & 7.928 & 5.838 & 4.839 \\
 & {\tiny 2,3} & -- & {\tiny 2} & {\tiny 1,2,3,5,6,7,8,9} & {\tiny 1,2,3,6,8,9} & {\tiny 1,2,3,9} & {\tiny 1,2,3,5,6,8,9} & {\tiny 1,2,3,6,9} & {\tiny 1,2,3} \\
SNR:0 & 2.808 & 1.593 & 1.796 & 8.399 & 6.281 & 5.196 & 7.926 & 6.017 & 4.985 \\
 & {\tiny 2,3} & -- & {\tiny 2} & {\tiny 1,2,3,5,6,7,8,9} & {\tiny 1,2,3,6,8,9} & {\tiny 1,2,3} & {\tiny 1,2,3,5,6,8,9} & {\tiny 1,2,3,6,9} & {\tiny 1,2,3} \\

\end{tabular}
\end{sidewaystable}
}

{
\setlength\tabcolsep{2.0pt}%
\begin{sidewaystable}[htb]
\centering
\scriptsize
\caption{Comparison of the attribute weighting methods for $F_1$ criterion. Average ranks and statistical tests.\label{table:Ex1_f1_stats}}
\begin{tabular}{l|lllllllll}
 & \multicolumn{9}{c}{$F_1$} \\

 & {\scriptsize B, KNN (1)} & {\scriptsize B, NBM (2)} & {\scriptsize B, GNB (3)} & {\scriptsize AW, KNN (4)} & {\scriptsize AW, NBM (5)} & {\scriptsize AW, GNB (6)} & {\scriptsize AWc, KNN (7)} & {\scriptsize AWc, NBM (8)} & {\scriptsize AWc, GNB (9)} \\
 \toprule

SNR:12 & 7.225 & 3.978 & 3.571 & 7.256 & 5.271 & 3.293 & 6.907 & 4.724 & 2.776 \\
 & {\tiny 2,3,5,6,7,8,9} & {\tiny 3,6,9} & {\tiny 6,9} & {\tiny 2,3,5,6,7,8,9} & {\tiny 2,3,6,8,9} & {\tiny 9} & {\tiny 2,3,5,6,8,9} & {\tiny 2,3,6,9} & -- \\
SNR:10 & 6.850 & 3.513 & 3.633 & 7.351 & 5.381 & 3.425 & 7.006 & 4.890 & 2.951 \\
 & {\tiny 2,3,5,6,8,9} & {\tiny 9} & {\tiny 9} & {\tiny 1,2,3,5,6,7,8,9} & {\tiny 2,3,6,8,9} & {\tiny 9} & {\tiny 2,3,5,6,8,9} & {\tiny 2,3,6,9} & -- \\
SNR:6 & 6.119 & 2.851 & 3.290 & 7.374 & 5.596 & 4.176 & 7.229 & 4.828 & 3.536 \\
 & {\tiny 2,3,5,6,8,9} & -- & {\tiny 2} & {\tiny 1,2,3,5,6,7,8,9} & {\tiny 2,3,6,8,9} & {\tiny 2,3,9} & {\tiny 1,2,3,5,6,8,9} & {\tiny 2,3,6,9} & {\tiny 2,3} \\
SNR:5 & 5.660 & 2.539 & 3.013 & 7.658 & 5.693 & 4.315 & 7.207 & 5.015 & 3.900 \\
 & {\tiny 2,3,6,8,9} & -- & {\tiny 2} & {\tiny 1,2,3,5,6,7,8,9} & {\tiny 1,2,3,6,8,9} & {\tiny 2,3,9} & {\tiny 1,2,3,5,6,8,9} & {\tiny 2,3,6,9} & {\tiny 2,3} \\
SNR:4 & 5.215 & 2.201 & 2.796 & 7.842 & 5.639 & 4.594 & 7.532 & 5.065 & 4.115 \\
 & {\tiny 2,3,6,8,9} & -- & {\tiny 2} & {\tiny 1,2,3,5,6,7,8,9} & {\tiny 2,3,6,8,9} & {\tiny 2,3,9} & {\tiny 1,2,3,5,6,8,9} & {\tiny 2,3,6,9} & {\tiny 2,3} \\
SNR:3 & 4.478 & 1.829 & 2.499 & 7.940 & 6.011 & 4.926 & 7.503 & 5.389 & 4.425 \\
 & {\tiny 2,3} & -- & {\tiny 2} & {\tiny 1,2,3,5,6,7,8,9} & {\tiny 1,2,3,6,8,9} & {\tiny 2,3,9} & {\tiny 1,2,3,5,6,8,9} & {\tiny 1,2,3,6,9} & {\tiny 2,3} \\
SNR:2 & 3.871 & 1.712 & 2.215 & 8.025 & 6.044 & 5.029 & 7.717 & 5.721 & 4.665 \\
 & {\tiny 2,3} & -- & {\tiny 2} & {\tiny 1,2,3,5,6,7,8,9} & {\tiny 1,2,3,6,8,9} & {\tiny 1,2,3,9} & {\tiny 1,2,3,5,6,8,9} & {\tiny 1,2,3,6,9} & {\tiny 1,2,3} \\
SNR:1 & 3.154 & 1.699 & 2.003 & 8.168 & 6.203 & 5.126 & 7.946 & 5.847 & 4.854 \\
 & {\tiny 2,3} & -- & {\tiny 2} & {\tiny 1,2,3,5,6,7,8,9} & {\tiny 1,2,3,6,8,9} & {\tiny 1,2,3,9} & {\tiny 1,2,3,5,6,8,9} & {\tiny 1,2,3,6,9} & {\tiny 1,2,3} \\
SNR:0 & 2.796 & 1.603 & 1.792 & 8.387 & 6.281 & 5.188 & 7.925 & 6.042 & 4.987 \\
 & {\tiny 2,3} & -- & {\tiny 2} & {\tiny 1,2,3,5,6,7,8,9} & {\tiny 1,2,3,6,8,9} & {\tiny 1,2,3} & {\tiny 1,2,3,5,6,8,9} & {\tiny 1,2,3,6,9} & {\tiny 1,2,3} \\

\end{tabular}
\end{sidewaystable}
}
\FloatBarrier

\subsection{Comparing the proposed method with different membership functions of the fuzzy set $T_l$}\label{sec:resultsanddisc:MembershipFunctions}

\begin{figure}[htb]
    \centering
        \includegraphics[width=0.99\columnwidth, height=.3\textheight, keepaspectratio]{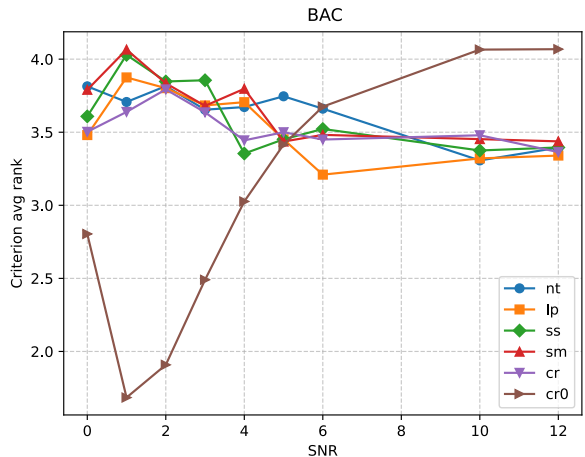}
    \caption{Comparison of the fuzzy membership functions. Average ranks plots for BAC criterion.}
    \label{figs:Ex2_bac}
\end{figure}

\begin{figure}[htb]
    \centering
        \includegraphics[width=0.99\columnwidth, height=.3\textheight, keepaspectratio]{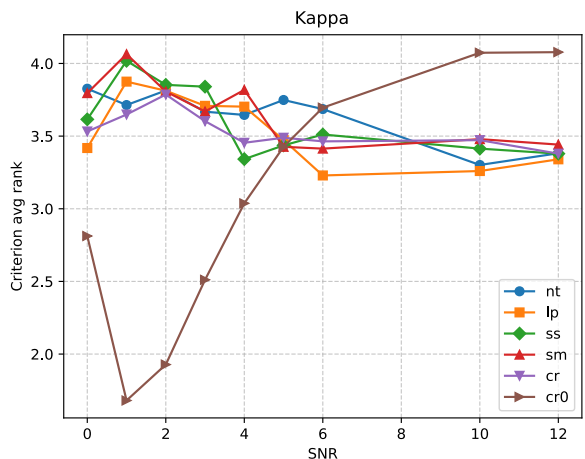}
    \caption{Comparison of the fuzzy membership functions. Average ranks plots for Kappa criterion.}
    \label{figs:Ex2_kappa}
\end{figure}

\begin{figure}[htb]
    \centering
        \includegraphics[width=0.99\columnwidth, height=.3\textheight, keepaspectratio]{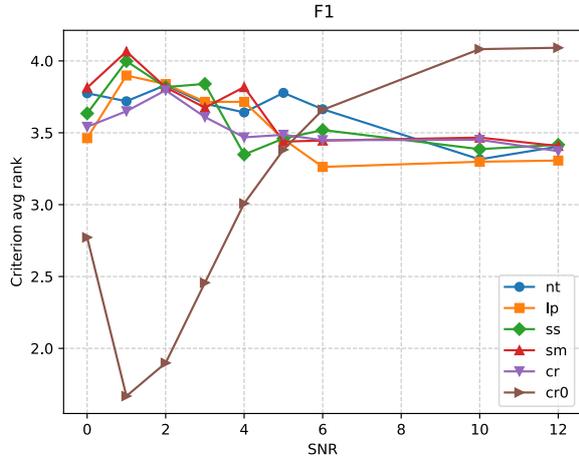}
    \caption{Comparison of the fuzzy membership functions. Average ranks plots for $F_1$ criterion.}
    \label{figs:Ex2_f1}
\end{figure}

In this experiment, we compare the performance of the proposed method with different membership functions of the fuzzy set $T_l$. The results related to this experiment are given in Figures~\ref{figs:Ex2_bac} -- \ref{figs:Ex2_f1} and Tables~\ref{table:Ex2_bac_stats} -- \ref{table:Ex2_f1_stats}. As in the section above, the quality-criteria-related results are also very consistent. Consequently, they are discussed together.

Let us start with the results related to \textbf{cr0} that always give 1.0 as the value of the membership function. We can see that this approach performs rather poorly for low SNR values but improves with the SNR value. This is expected as the membership function does not take into account the contamination level of the signal. For $\mathrm{SNR} \in \{10,12\}$, the \textbf{cr0} method is significantly better than the remaining membership functions. This is because at high SNR levels it can be safely assumed that signals contain no contaminations. However, for $\mathrm{SNR} \leq 4$ the remaning membership functions allow the attribute-weighted KNN classifier to achieve significanlty better classification quality. There are also almost no significant differences between the other membership functions. The only membership creation functions that, for some SNR values, perform significantly better than other functions are \textbf{nt} and \textbf{sm}. However, these differences, despite their statistical significance, are rather inconsistent and do not show a clear trend. The impact of the membership function on the classification quality is therefore rather small in the case of the datasets used in this study.

{
\setlength\tabcolsep{2.0pt}%
\begin{table}[htb]
\centering
\scriptsize
\caption{Comparison of the fuzzy membership functions for BAC criterion. Average ranks and statistical tests.\label{table:Ex2_bac_stats}}
\begin{tabular}{l|llllll}
 & \multicolumn{6}{c}{BAC} \\
 & nt (1) & lp (2) & ss (3) & sm (4) & cr (5) & cr0 (6) \\
 \toprule 
 SNR:12 & 3.394 & 3.340 & 3.396 & 3.438 & 3.364 & 4.068 \\
 & -- & -- & -- & -- & -- & {\tiny 1,2,3,4,5} \\
SNR:10 & 3.307 & 3.321 & 3.375 & 3.453 & 3.479 & 4.065 \\
 & -- & -- & -- & -- & -- & {\tiny 1,2,3,4,5} \\
SNR:6 & 3.661 & 3.210 & 3.522 & 3.482 & 3.450 & 3.675 \\
 & {\tiny 2} & -- & -- & -- & -- & {\tiny 2} \\
SNR:5 & 3.746 & 3.451 & 3.451 & 3.436 & 3.500 & 3.415 \\
 & -- & -- & -- & -- & -- & -- \\
SNR:4 & 3.672 & 3.706 & 3.354 & 3.797 & 3.444 & 3.026 \\
 & {\tiny 6} & {\tiny 6} & {\tiny 6} & {\tiny 3,5,6} & {\tiny 6} & -- \\
SNR:3 & 3.654 & 3.683 & 3.856 & 3.681 & 3.636 & 2.490 \\
 & {\tiny 6} & {\tiny 6} & {\tiny 6} & {\tiny 6} & {\tiny 6} & -- \\
SNR:2 & 3.815 & 3.801 & 3.847 & 3.835 & 3.793 & 1.908 \\
 & {\tiny 6} & {\tiny 6} & {\tiny 6} & {\tiny 6} & {\tiny 6} & -- \\
SNR:1 & 3.707 & 3.875 & 4.029 & 4.065 & 3.639 & 1.685 \\
 & {\tiny 6} & {\tiny 6} & {\tiny 6} & {\tiny 6} & {\tiny 6} & -- \\
SNR:0 & 3.814 & 3.481 & 3.608 & 3.790 & 3.503 & 2.804 \\
 & {\tiny 6} & {\tiny 6} & {\tiny 6} & {\tiny 6} & {\tiny 6} & -- \\

\end{tabular}

\end{table}
}

{
\setlength\tabcolsep{2.0pt}%
\begin{table}[htb]
\centering
\scriptsize
\caption{Comparison of the fuzzy membership functions for Kappa criterion. Average ranks and statistical tests.\label{table:Ex2_kappa_stats}}
\begin{tabular}{l|llllll}
 & \multicolumn{6}{c}{Kappa} \\
 & nt (1) & lp (2) & ss (3) & sm (4) & cr (5) & cr0 (6) \\
 \toprule 
 SNR:12 & 3.382 & 3.340 & 3.379 & 3.442 & 3.379 & 4.078 \\
 & -- & -- & -- & -- & -- & {\tiny 1,2,3,4,5} \\
SNR:10 & 3.301 & 3.260 & 3.414 & 3.479 & 3.472 & 4.074 \\
 & -- & -- & -- & -- & -- & {\tiny 1,2,3,4,5} \\
SNR:6 & 3.686 & 3.229 & 3.511 & 3.414 & 3.464 & 3.696 \\
 & {\tiny 2} & -- & -- & -- & -- & {\tiny 2} \\
SNR:5 & 3.749 & 3.474 & 3.436 & 3.426 & 3.489 & 3.426 \\
 & -- & -- & -- & -- & -- & -- \\
SNR:4 & 3.646 & 3.703 & 3.342 & 3.818 & 3.454 & 3.038 \\
 & {\tiny 6} & {\tiny 6} & {\tiny 6} & {\tiny 3,6} & {\tiny 6} & -- \\
SNR:3 & 3.668 & 3.708 & 3.840 & 3.669 & 3.603 & 2.511 \\
 & {\tiny 6} & {\tiny 6} & {\tiny 6} & {\tiny 6} & {\tiny 6} & -- \\
SNR:2 & 3.814 & 3.812 & 3.853 & 3.807 & 3.786 & 1.928 \\
 & {\tiny 6} & {\tiny 6} & {\tiny 6} & {\tiny 6} & {\tiny 6} & -- \\
SNR:1 & 3.714 & 3.875 & 4.018 & 4.064 & 3.649 & 1.681 \\
 & {\tiny 6} & {\tiny 6} & {\tiny 5,6} & {\tiny 6} & {\tiny 6} & -- \\
SNR:0 & 3.826 & 3.418 & 3.615 & 3.796 & 3.532 & 2.812 \\
 & {\tiny 6} & {\tiny 6} & {\tiny 6} & {\tiny 6} & {\tiny 6} & -- \\

\end{tabular}

\end{table}
}

{
\setlength\tabcolsep{2.0pt}%
\begin{table}[htb]
\centering
\scriptsize
\caption{Comparison of the fuzzy membership functions for $F_1$ criterion. Average ranks and statistical tests.\label{table:Ex2_f1_stats}}
\begin{tabular}{l|llllll}
 & \multicolumn{6}{c}{$F_1$} \\
 & nt (1) & lp (2) & ss (3) & sm (4) & cr (5) & cr0 (6) \\
 \toprule

 SNR:12 & 3.404 & 3.307 & 3.417 & 3.408 & 3.372 & 4.092 \\
 & -- & -- & -- & -- & -- & {\tiny 1,2,3,4,5} \\
SNR:10 & 3.315 & 3.299 & 3.386 & 3.467 & 3.451 & 4.082 \\
 & -- & -- & -- & -- & -- & {\tiny 1,2,3,4,5} \\
SNR:6 & 3.664 & 3.263 & 3.519 & 3.446 & 3.449 & 3.660 \\
 & {\tiny 2} & -- & -- & -- & -- & {\tiny 2} \\
SNR:5 & 3.778 & 3.456 & 3.460 & 3.439 & 3.486 & 3.382 \\
 & -- & -- & -- & -- & -- & -- \\
SNR:4 & 3.642 & 3.715 & 3.349 & 3.818 & 3.468 & 3.008 \\
 & {\tiny 6} & {\tiny 6} & {\tiny 6} & {\tiny 3,6} & {\tiny 6} & -- \\
SNR:3 & 3.703 & 3.715 & 3.840 & 3.675 & 3.610 & 2.457 \\
 & {\tiny 6} & {\tiny 6} & {\tiny 6} & {\tiny 6} & {\tiny 6} & -- \\
SNR:2 & 3.832 & 3.839 & 3.818 & 3.817 & 3.796 & 1.899 \\
 & {\tiny 6} & {\tiny 6} & {\tiny 6} & {\tiny 6} & {\tiny 6} & -- \\
SNR:1 & 3.719 & 3.899 & 3.999 & 4.065 & 3.650 & 1.668 \\
 & {\tiny 6} & {\tiny 6} & {\tiny 6} & {\tiny 6} & {\tiny 6} & -- \\
SNR:0 & 3.775 & 3.462 & 3.635 & 3.814 & 3.542 & 2.772 \\
 & {\tiny 6} & {\tiny 6} & {\tiny 6} & {\tiny 6} & {\tiny 6} & -- \\

\end{tabular}

\end{table}
}
\FloatBarrier

\subsection{Comparing the proposed method with reference methods taken from the literature}\label{sec:resultsanddisc:ReferenceMethods}

\begin{figure}[htb]
    \centering
        \includegraphics[width=0.99\columnwidth, height=.3\textheight, keepaspectratio]{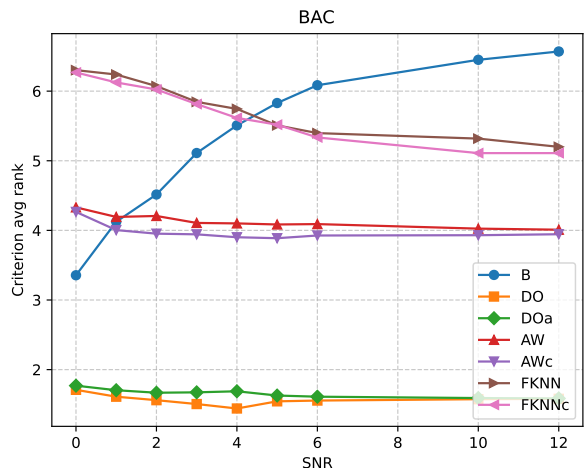}
    \caption{Comparison with the literature methods. Average ranks for Balanced Accuracy Criterion.}
    \label{figs:Ex3_bac}
\end{figure}

\begin{figure}[htb]
    \centering
        \includegraphics[width=0.99\columnwidth, height=.3\textheight, keepaspectratio]{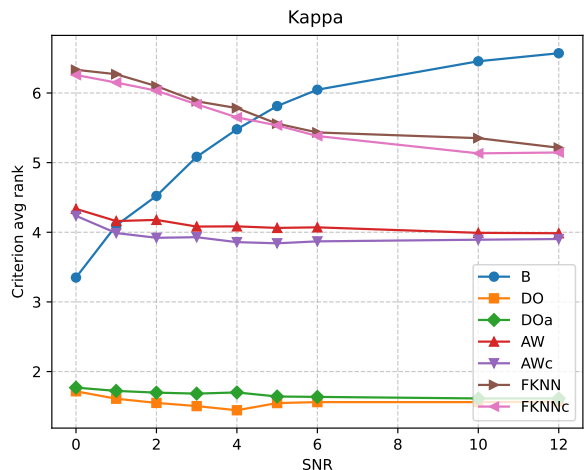}
    \caption{Comparison with the literature methods. Average ranks for Kappa Criterion.}
    \label{figs:Ex3_kappa}
\end{figure}

\begin{figure}[htb]
    \centering
        \includegraphics[width=0.99\columnwidth, height=.3\textheight, keepaspectratio]{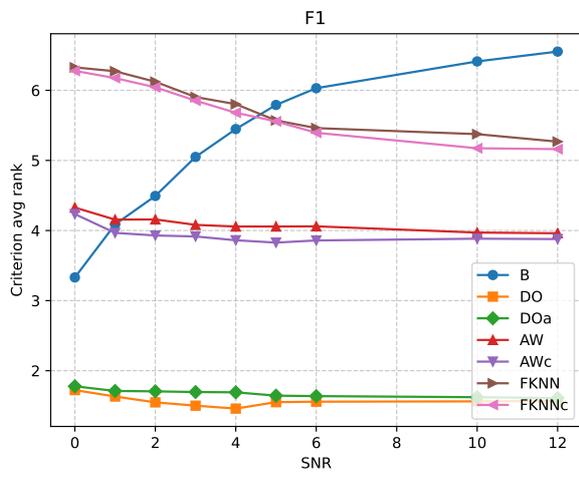}
    \caption{Comparison with the literature methods. Average ranks for $F_1$ Criterion.}
    \label{figs:Ex3_f1}
\end{figure}

In this experiment, we compare the proposed method with reference methods taken from the literature. The results related to this comparison are given in Figures~\ref{figs:Ex3_bac} -- \ref{figs:Ex3_f1} and Tables~\ref{table:Ex3_bac_stats} -- \ref{table:Ex3_f1_stats}.

The results for the \textbf{B} method show how a single KNN classifier, unaware of signal contamination, would perform at different SNR values. The results shown allow the reader to clearly distinguish between two groups of methods: those based on a weighted classifier combination at the level of soft predictions (\textbf{DO}, \textbf{DOa}) and those that incorporate noise-related information at different levels of the classification process (\textbf{AW}, \textbf{AWc}, \textbf{FKNN}, \textbf{FKNNc}). Among the latter methods, there are also two groups of approaches: those based on a single classifier that incorporates attribute weighting (\textbf{AW}, \textbf{AWc}) and those that incorporate noise-related information in the fuzzy set-based classifier combination (\textbf{FKNN}, \textbf{FKNNc}).

The base method \textbf{B} that do not use information about channel contamination achieves the best classification quality for high SNR values. For $\mathrm{SNR} \geq 5$, it outperforms other methods. 

The first group of methods (\textbf{DO}, \textbf{DOa}) performs rather poorly for all SNR values. For all SNR values, the performance of these methods is significantly worse than the performance of the other methods investigated. This confirms the observation from~\cite{Trajdos2025} that single-channel-based classifiers, combined at the soft support level, are unable to fully utilise the cross-channel information present in the EMG signals. Consequently, the performance of these methods is rather poor compared to that of other methods investigated in this paper.
 
 The second group of methods (\textbf{AW}, \textbf{AWc}, \textbf{FKNN}, \textbf{FKNNc}) performs significantly better for all SNR values than the methods belonging to the first group. They also outperform \textbf{B} for low SNR values.
 
 The \textbf{FKNN} and \textbf{FKNNc} significantly outperform \textbf{AW} and \textbf{AWc} for all SNR values. It shows that fuzzy-set-based strategy tends to be better than simple attribute weighting approaches. The results also show that there are no significant differences when \textbf{FKNN} and \textbf{FKNNc} are compared. This result confirms the observation made in the previous section that the choice of the fuzzy membership function does not make a significant difference for the \textbf{FKNN}-based method unless the membership function can distinguish between contaminated and uncontaminated samples.

{
\setlength\tabcolsep{2.0pt}%

\begin{table*}[htb]
\centering
\scriptsize
\caption{Comparison with the literature methods for BAC criterion. Average ranks and statistical tests.\label{table:Ex3_bac_stats}}
\begin{tabular}{l|lllllll}
 & \multicolumn{7}{c}{BAC} \\
 & B (1) & DO (2) & DOa (3) & AW  (4) & AWc (5) & FKNN (6) & FKNNc (7) \\
\toprule
SNR:12 & 6.569 & 1.581 & 1.586 & 4.010 & 3.944 & 5.200 & 5.110 \\
 & {\tiny 2,3,4,5,6,7} & -- & -- & {\tiny 2,3} & {\tiny 2,3} & {\tiny 2,3,4,5} & {\tiny 2,3,4,5} \\
SNR:10 & 6.450 & 1.575 & 1.592 & 4.025 & 3.931 & 5.318 & 5.110 \\
 & {\tiny 2,3,4,5,6,7} & -- & -- & {\tiny 2,3} & {\tiny 2,3} & {\tiny 2,3,4,5} & {\tiny 2,3,4,5} \\
SNR:6 & 6.085 & 1.556 & 1.611 & 4.090 & 3.926 & 5.397 & 5.335 \\
 & {\tiny 2,3,4,5,6,7} & -- & -- & {\tiny 2,3,5} & {\tiny 2,3} & {\tiny 2,3,4,5} & {\tiny 2,3,4,5} \\
SNR:5 & 5.829 & 1.544 & 1.628 & 4.085 & 3.888 & 5.508 & 5.518 \\
 & {\tiny 2,3,4,5,6,7} & -- & {\tiny 2} & {\tiny 2,3,5} & {\tiny 2,3} & {\tiny 2,3,4,5} & {\tiny 2,3,4,5} \\
SNR:4 & 5.508 & 1.442 & 1.689 & 4.100 & 3.901 & 5.744 & 5.615 \\
 & {\tiny 2,3,4,5} & -- & {\tiny 2} & {\tiny 2,3,5} & {\tiny 2,3} & {\tiny 2,3,4,5} & {\tiny 2,3,4,5} \\
SNR:3 & 5.111 & 1.506 & 1.674 & 4.107 & 3.943 & 5.846 & 5.814 \\
 & {\tiny 2,3,4,5} & -- & {\tiny 2} & {\tiny 2,3} & {\tiny 2,3} & {\tiny 1,2,3,4,5} & {\tiny 1,2,3,4,5} \\
SNR:2 & 4.515 & 1.561 & 1.668 & 4.207 & 3.953 & 6.072 & 6.024 \\
 & {\tiny 2,3,4,5} & -- & {\tiny 2} & {\tiny 2,3,5} & {\tiny 2,3} & {\tiny 1,2,3,4,5} & {\tiny 1,2,3,4,5} \\
SNR:1 & 4.124 & 1.611 & 1.704 & 4.193 & 4.004 & 6.239 & 6.125 \\
 & {\tiny 2,3,5} & -- & -- & {\tiny 1,2,3,5} & {\tiny 2,3} & {\tiny 1,2,3,4,5} & {\tiny 1,2,3,4,5} \\
SNR:0 & 3.356 & 1.710 & 1.769 & 4.331 & 4.265 & 6.301 & 6.268 \\
 & {\tiny 2,3} & -- & -- & {\tiny 1,2,3} & {\tiny 1,2,3} & {\tiny 1,2,3,4,5} & {\tiny 1,2,3,4,5} \\
\end{tabular}
\end{table*}
}

{
\setlength\tabcolsep{2.0pt}%

\begin{table*}[htb]
\centering
\scriptsize
\caption{Comparison with the literature methods for Kappa criterion. Average ranks and statistical tests.\label{table:Ex3_kappa_stats}}
\begin{tabular}{l|lllllll}
 & \multicolumn{7}{c}{Kappa} \\
 & B (1) & DO (2) & DOa (3) & AW  (4) & AWc (5) & FKNN (6) & FKNNc (7) \\
\toprule
SNR:12 & 6.571 & 1.567 & 1.614 & 3.986 & 3.903 & 5.214 & 5.146 \\
 & {\tiny 2,3,4,5,6,7} & -- & -- & {\tiny 2,3} & {\tiny 2,3} & {\tiny 2,3,4,5} & {\tiny 2,3,4,5} \\
SNR:10 & 6.456 & 1.561 & 1.614 & 3.992 & 3.893 & 5.351 & 5.133 \\
 & {\tiny 2,3,4,5,6,7} & -- & -- & {\tiny 2,3} & {\tiny 2,3} & {\tiny 2,3,4,5} & {\tiny 2,3,4,5} \\
SNR:6 & 6.047 & 1.561 & 1.636 & 4.071 & 3.869 & 5.433 & 5.382 \\
 & {\tiny 2,3,4,5,6,7} & -- & -- & {\tiny 2,3,5} & {\tiny 2,3} & {\tiny 2,3,4,5} & {\tiny 2,3,4,5} \\
SNR:5 & 5.812 & 1.547 & 1.642 & 4.062 & 3.842 & 5.560 & 5.535 \\
 & {\tiny 2,3,4,5,6,7} & -- & {\tiny 2} & {\tiny 2,3,5} & {\tiny 2,3} & {\tiny 2,3,4,5} & {\tiny 2,3,4,5} \\
SNR:4 & 5.479 & 1.444 & 1.700 & 4.085 & 3.858 & 5.783 & 5.650 \\
 & {\tiny 2,3,4,5} & -- & {\tiny 2} & {\tiny 2,3,5} & {\tiny 2,3} & {\tiny 1,2,3,4,5} & {\tiny 1,2,3,4,5} \\
SNR:3 & 5.083 & 1.504 & 1.683 & 4.082 & 3.929 & 5.881 & 5.838 \\
 & {\tiny 2,3,4,5} & -- & {\tiny 2} & {\tiny 2,3} & {\tiny 2,3} & {\tiny 1,2,3,4,5} & {\tiny 1,2,3,4,5} \\
SNR:2 & 4.522 & 1.550 & 1.697 & 4.178 & 3.921 & 6.099 & 6.033 \\
 & {\tiny 2,3,4,5} & -- & {\tiny 2} & {\tiny 2,3,5} & {\tiny 2,3} & {\tiny 1,2,3,4,5} & {\tiny 1,2,3,4,5} \\
SNR:1 & 4.099 & 1.610 & 1.722 & 4.161 & 3.989 & 6.269 & 6.150 \\
 & {\tiny 2,3,5} & -- & -- & {\tiny 1,2,3,5} & {\tiny 2,3} & {\tiny 1,2,3,4,5} & {\tiny 1,2,3,4,5} \\
SNR:0 & 3.350 & 1.717 & 1.769 & 4.336 & 4.236 & 6.333 & 6.258 \\
 & {\tiny 2,3} & -- & -- & {\tiny 1,2,3} & {\tiny 1,2,3} & {\tiny 1,2,3,4,5} & {\tiny 1,2,3,4,5} \\
\end{tabular}
\end{table*}
}

{
\setlength\tabcolsep{2.0pt}%

\begin{table*}[htb]
\centering
\scriptsize
\caption{Comparison with the literature methods for $F_1$ criterion. Average ranks and statistical tests.\label{table:Ex3_f1_stats}}
\begin{tabular}{l|lllllll}
 & \multicolumn{7}{c}{$F_1$} \\
 & B (1) & DO (2) & DOa (3) & AW  (4) & AWc (5) & FKNN (6) & FKNNc (7) \\
\toprule
SNR:12 & 6.553 & 1.569 & 1.610 & 3.961 & 3.878 & 5.268 & 5.161 \\
 & {\tiny 2,3,4,5,6,7} & -- & -- & {\tiny 2,3} & {\tiny 2,3} & {\tiny 2,3,4,5} & {\tiny 2,3,4,5} \\
SNR:10 & 6.412 & 1.562 & 1.622 & 3.971 & 3.883 & 5.375 & 5.174 \\
 & {\tiny 2,3,4,5,6,7} & -- & -- & {\tiny 2,3} & {\tiny 2,3} & {\tiny 2,3,4,5} & {\tiny 2,3,4,5} \\
SNR:6 & 6.031 & 1.558 & 1.637 & 4.060 & 3.858 & 5.461 & 5.394 \\
 & {\tiny 2,3,4,5,6,7} & -- & -- & {\tiny 2,3,5} & {\tiny 2,3} & {\tiny 2,3,4,5} & {\tiny 2,3,4,5} \\
SNR:5 & 5.792 & 1.551 & 1.644 & 4.057 & 3.828 & 5.571 & 5.557 \\
 & {\tiny 2,3,4,5,6,7} & -- & {\tiny 2} & {\tiny 2,3,5} & {\tiny 2,3} & {\tiny 2,3,4,5} & {\tiny 2,3,4,5} \\
SNR:4 & 5.447 & 1.460 & 1.692 & 4.057 & 3.861 & 5.803 & 5.681 \\
 & {\tiny 2,3,4,5} & -- & {\tiny 2} & {\tiny 2,3,5} & {\tiny 2,3} & {\tiny 1,2,3,4,5} & {\tiny 1,2,3,4,5} \\
SNR:3 & 5.050 & 1.501 & 1.696 & 4.081 & 3.915 & 5.904 & 5.853 \\
 & {\tiny 2,3,4,5} & -- & {\tiny 2} & {\tiny 2,3} & {\tiny 2,3} & {\tiny 1,2,3,4,5} & {\tiny 1,2,3,4,5} \\
SNR:2 & 4.493 & 1.549 & 1.706 & 4.158 & 3.929 & 6.122 & 6.043 \\
 & {\tiny 2,3,4,5} & -- & {\tiny 2} & {\tiny 2,3,5} & {\tiny 2,3} & {\tiny 1,2,3,4,5} & {\tiny 1,2,3,4,5} \\
SNR:1 & 4.088 & 1.632 & 1.711 & 4.156 & 3.967 & 6.272 & 6.175 \\
 & {\tiny 2,3,5} & -- & -- & {\tiny 1,2,3,5} & {\tiny 2,3} & {\tiny 1,2,3,4,5} & {\tiny 1,2,3,4,5} \\
SNR:0 & 3.331 & 1.724 & 1.778 & 4.328 & 4.235 & 6.328 & 6.278 \\
 & {\tiny 2,3} & -- & -- & {\tiny 1,2,3} & {\tiny 1,2,3} & {\tiny 1,2,3,4,5} & {\tiny 1,2,3,4,5} \\
\end{tabular}
\end{table*}
}
\FloatBarrier
\section{Conclusions}\label{sec:Conclusions}

In this paper, we propose a new recognition system intended for the electromyographic control of a bionic upper limb prosthesis with the detection of contaminated biosignals to mitigate the adverse effect of contamination and improve the quality of the classification. The proposed recognition system includes two multiclassifiers. The first is an outlier detector based on one-class classification. The second is an ensemble of fuzzy KNN classifiers combined at the fuzzy set level. Experimental investigations were carried out using real biosignal signals from nine subjects with amputation.

The answers to the research questions posed in Section~\ref{sec:Introduction} are as follows:
\begin{enumerate}
    \item \ref{itm:rq:ent_feat_space} The proposed method significantly improves classification quality compared to a single noise-insensitive classifier trained on the entire feature space when the SNR levels are low.
    \item \ref{itm:rq:ltr_comp} The proposed method performs significantly better than the reference methods based on a single classifier combined at the soft support level. Both variants of the proposed method also outperform the attribute weighting/selection methods based on a single classifier.
    \item \ref{itm:rq:snr_vals}. The proposed method tends to be more robust to noise than the other methods when low SNR values are considered.
    \item \ref{itm:rq:fmember} For the investigated datasets, the membership function of the fuzzy set $T_l$ does not significantly affect the classification quality.
\end{enumerate}

The experimental study conducted confirms that the proposed method is capable of mitigate the adverse effect of contaminated biosignals and improve the quality of the classification. 

\FloatBarrier
% \clearpage
% \twocolumn[
% \vspace*{-.1cm} % Optional tweak if needed to push closer to top
% ]
\bibliography{bibliography}

\end{document}